\documentclass[lettersize,journal]{IEEEtran}

\usepackage{amsmath,amsfonts}
\usepackage{array}

\usepackage{algorithm}
\usepackage{algorithmic}

\usepackage{graphicx}
\usepackage[caption=false,font=normalsize,labelfont=sf,textfont=sf]{subfig}
\usepackage{booktabs}
\usepackage{multirow}
\usepackage{stfloats}

\usepackage{textcomp}
\usepackage{verbatim}
\usepackage{color}
\usepackage{epstopdf}
\usepackage{balance}

\usepackage[OT1]{fontenc}
\usepackage{mathptmx}
\usepackage[english]{babel}

\usepackage{cite}
\usepackage{url}

\usepackage{hyperref}
\usepackage{pifont}
\usepackage[mathscr]{eucal}

\def\BibTeX{{\rm B\kern-.05em{\sc i\kern-.025em b}\kern-.08em
		T\kern-.1667em\lower.7ex\hbox{E}\kern-.125emX}}

\begin{document}

\title{UniPlanner: A Unified Motion Planning Framework for Autonomous Vehicle Decision-Making Systems via Multi-Dataset Integration	
}

\author{Xin Yang, Yuhang Zhang, Wei Li, Xin Lin, Wenbin Zou, Chen Xu
\thanks{Xin Yang, Wei Li, Wenbin Zou, and Chen Xu  are with the Guangdong Key Laboratory of Intelligent Information Processing, College of Electronics and Information Engineering, Shenzhen University, Shenzhen 518000, China (e-mail: 2350432002@email.szu.edu.cn;
	 2250432002@email.szu.edu.cn; wzou@szu.edu.cn;xuchen@szu.edu.cn;)
	}
\thanks{Yuhang Zhang is with the School of Computer Science and Cyber Engineering, Guangzhou University, Guangzhou, China(e-mail:yuhang.zhang@gzhu.edu.cn)
}
\thanks{Xin Lin is with the School of Artificial Intelligence, Guangzhou University, Guangzhou, China (e-mail: linxin94@gzhu.edu.cn).
}
\thanks{(Corresponding author: Wenbin Zou.)}}

\markboth{Journal of \LaTeX\ Class Files,~Vol.~14, No.~8, August~2021}%
{Shell \MakeLowercase{\textit{et al.}}: A Sample Article Using IEEEtran.cls for IEEE Journals}

\maketitle

\begin{abstract}

	Motion planning is a critical component of autonomous vehicle decision-making systems, directly determining trajectory safety and driving efficiency. While deep learning approaches have advanced planning capabilities, existing methods remain confined to single-dataset training, limiting their robustness in planning.
	Through systematic analysis, we discover that vehicular trajectory distributions and history-future correlations demonstrate remarkable consistency across different datasets. Based on these findings, we propose UniPlanner, the first planning framework designed for multi-dataset integration in autonomous vehicle decision-making. UniPlanner achieves unified cross-dataset learning through three synergistic innovations.
	First, the History-Future Trajectory Dictionary Network (HFTDN) aggregates history-future trajectory pairs from multiple datasets, using historical trajectory similarity to retrieve relevant futures and generate cross-dataset planning guidance.
	Second, the Gradient-Free Trajectory Mapper (GFTM) learns robust history-future correlations from multiple datasets, transforming historical trajectories into universal planning priors. Its gradient-free design ensures the introduction of valuable priors while preventing shortcut learning, making the planning knowledge safely transferable. Third, the Sparse-to-Dense (S2D) paradigm implements adaptive dropout to selectively suppress planning priors during training for robust learning, while enabling full prior utilization during inference to maximize planning performance.
	Extensive experiments and ablation studies confirm UniPlanner's ability to achieve significant performance gains through multi-dataset integration. Ablations particularly demonstrate that leveraging universal trajectory correlations across datasets drives these improvements, establishing a new paradigm for multi-dataset motion planning. Code will be released upon publication at \url{https://github.com/942411526/UniPlanner}.

\end{abstract}

\section{Introduction}
\IEEEPARstart{M}{otion} planning serves as the core decision-making module in Connected and Autonomous Vehicles (CAVs), directly controlling vehicle trajectories and ensuring navigation safety in complex traffic environments \cite{ref41,ref49,ref50}. Recent end-to-end approaches have achieved remarkable performance through imitation learning \cite{ref45}, reinforcement learning \cite{ref44} \cite{ref52}, and hybrid prediction-planning architectures \cite{ref3}, successfully handling challenging scenarios including unprotected turns and dense traffic \cite{ref51}.

\begin{figure}[!t]
	\centering
	\includegraphics[width=3.5 in]{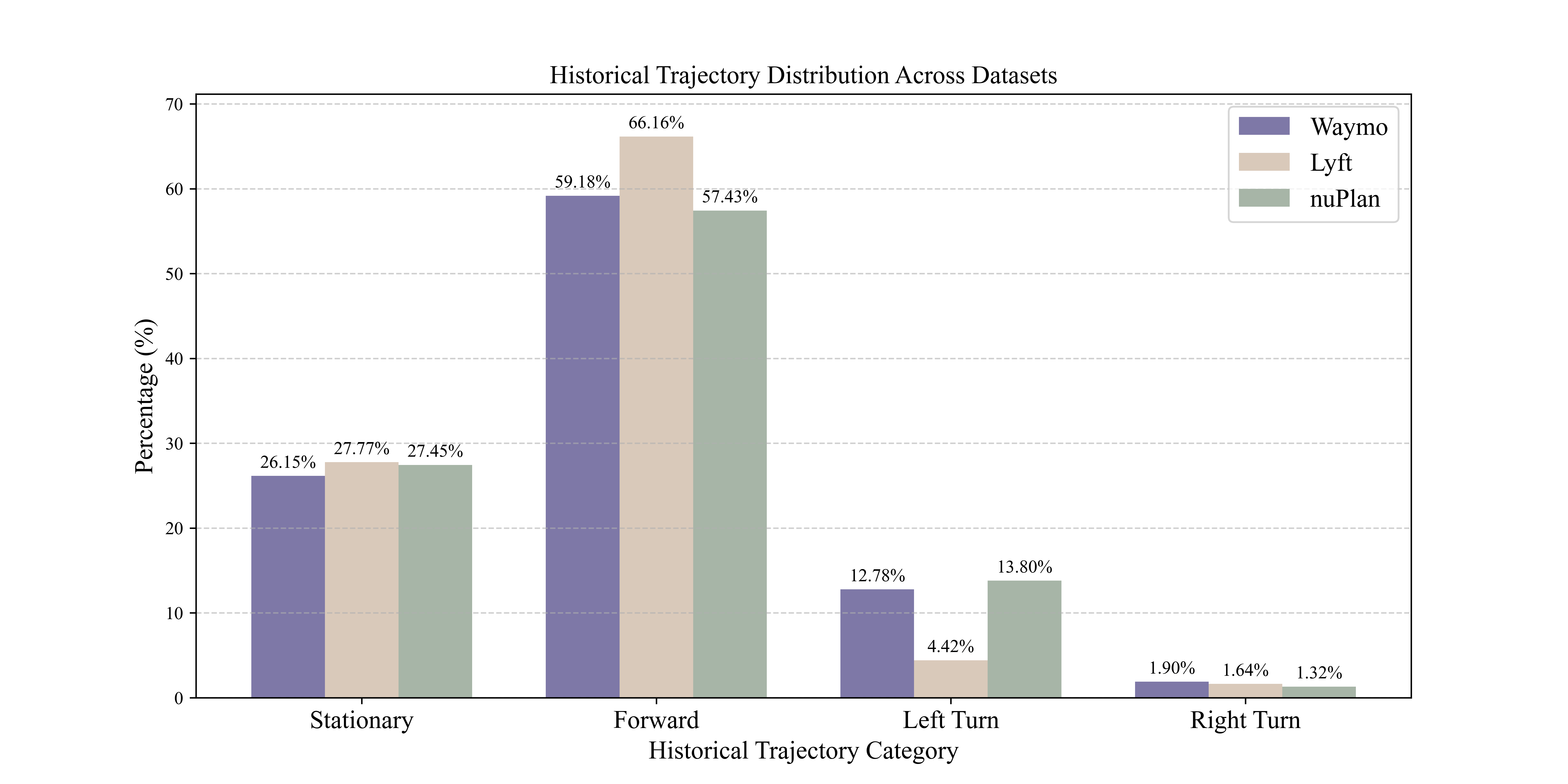}
	\caption{Motion type distribution of historical trajectories in Waymo \cite{ref19}, Lyft \cite{ref18}, and nuPlan \cite{ref17}}
	\label{fig:1}
\end{figure}

Despite these advances, current planning models remain vulnerable to distribution shifts and long-tail scenarios \cite{ref26}, limiting their deployment in autonomous vehicles across diverse traffic conditions.
Multi-dataset integration could enhance model robustness by expanding scenario coverage \cite{ref37,ref47,ref46}. \textbf{However, this paradigm remains unexplored in vehicle motion planning.}  Existing datasets exhibit severe heterogeneity in sensor configurations, coordinate systems, and annotation formats, hindering cross-dataset learning \cite{ref25}.
\textbf{Our analysis reveals that vehicle historical trajectories form a universal representation invariant across datasets.} Unlike sensor-specific measurements, trajectories reflect fundamental motion constraints and traffic patterns that transcend platform differences, enabling effective cross-dataset learning.

To validate this hypothesis, we analyze three large-scale autonomous driving datasets: Waymo \cite{ref19}, Lyft \cite{ref18}, and nuPlan \cite{ref17}, collectively representing over 2500 hours of real-world driving data. Using vehicle dynamics-based classification with \textit{curvature} and \textit{yaw rate}, we categorize trajectories into four fundamental maneuvers: \textit{Stationary}, \textit{Forward}, \textit{Left Turn}, and \textit{Right Turn}. Figure~\ref{fig:1} shows remarkably consistent trajectory distributions across all datasets, confirming that motion distributions remain invariant despite different sensor configurations and geographical locations. This consistency stems from road infrastructure constraints: straight segments and stops exhibit high frequency while turns remain relatively rare at intersections, producing consistent distributions across all datasets.

\begin{figure}[!t]
	\centering
	\includegraphics[width=3 in]{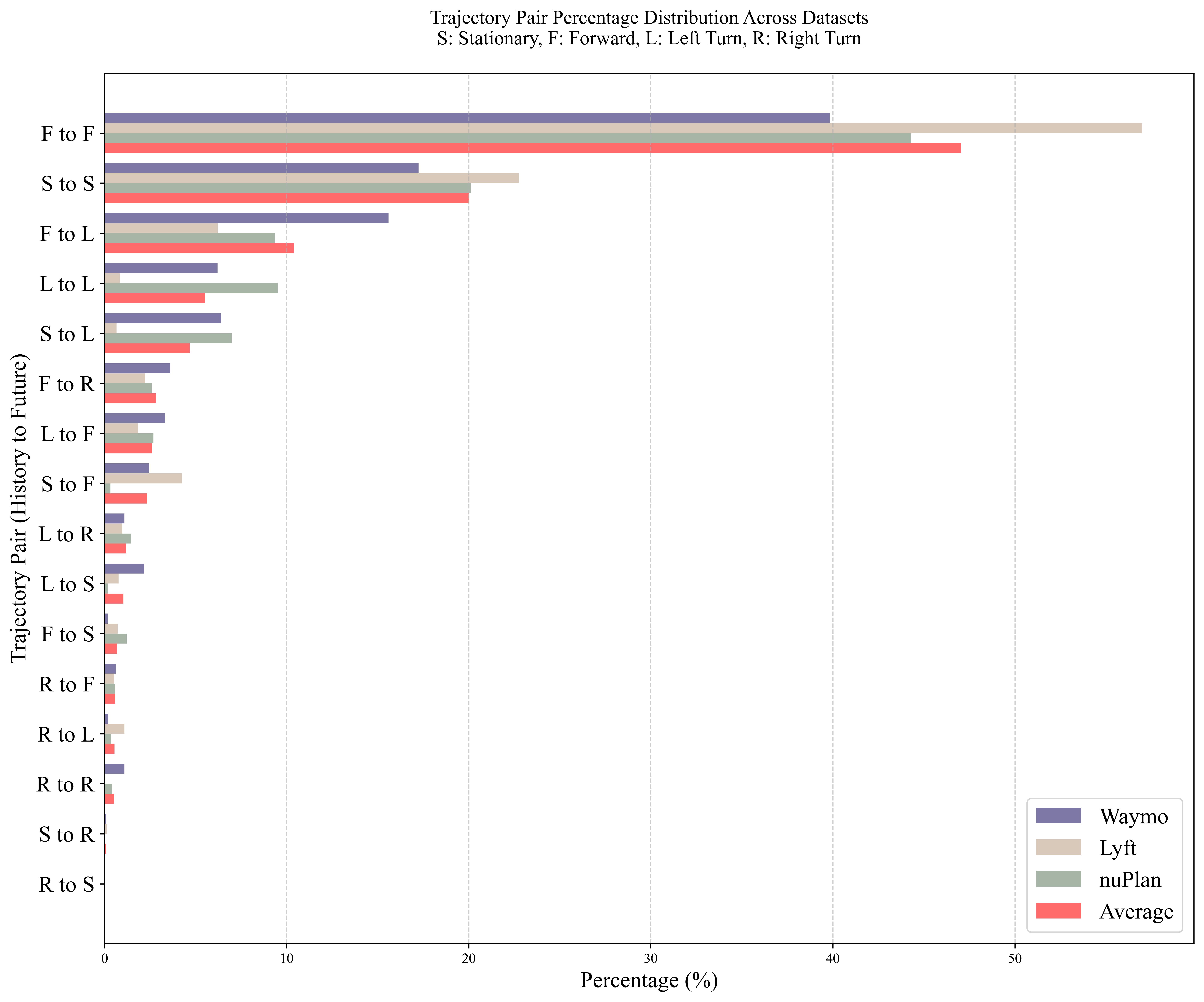}
	\caption{Statistical distribution of correlations between historical and future trajectory motion types across Waymo \cite{ref19}, Lyft \cite{ref18}, and nuPlan \cite{ref17}}
	\label{fig:2}
\end{figure}
Building upon the correlations identified in \cite{ref22}\cite{ref24}, we conduct systematic statistical analysis to validate and extend these findings across multiple datasets (Figure~\ref{fig:2}). Our analysis reveals remarkable universality in temporal correlations of driving behavior across diverse datasets. Specifically, seven key transition patterns: $\textit{Forward} \rightarrow \textit{Forward}$, $\textit{Stationary} \rightarrow \textit{Stationary}$, $\textit{Forward} \rightarrow \textit{Left Turn}$, $\textit{Left Turn} \rightarrow \textit{Left Turn}$, $\textit{Stationary} \rightarrow \textit{Left Turn}$, $\textit{Forward} \rightarrow \textit{Right Turn}$, and $\textit{Left Turn} \rightarrow \textit{Forward}$, consistently dominate across all examined datasets, collectively accounting for over 80\% of observed transitions.
Quantitative evaluation reveals that history-future trajectory correlations maintain remarkable consistency across diverse driving datasets. Through extensive cross-dataset analysis, we identify two fundamental properties:

\textbf{(i) Distributional Invariance:} Historical trajectory representations exhibit similar statistical distributions across different datasets.

\textbf{(ii) Temporal Correlation Consistency:} History-future temporal correlations demonstrate consistent across different datasets.

 Based on these findings, we propose UniPlanner, the first framework achieving multi-dataset integration for autonomous vehicle motion planning through two key innovations: (1) gradient-free trajectory mapping that captures dataset-invariant history-future correlations from multiple datasets to generate planning priors, with gradient isolation preventing harmful dependencies; (2) a cross-dataset trajectory dictionary that retrieves candidate trajectories via historical similarity and temporal correlations, transforming multi-dataset experiences into actionable planning guidance. The main contributions of this paper are summarized as follows:
\begin{itemize}
	\item We conduct the first systematic analysis of trajectory properties across multiple autonomous driving datasets, revealing universal distributions and history-future correlations that enable cross-dataset learning. Based on these invariances, UniPlanner pioneers multi-dataset motion planning, proving that cross-dataset learning can enhance autonomous vehicle performance.
	
	\item We propose the Gradient-Free Trajectory Mapper (GFTM), which learns universal history-future correlations from multiple datasets while preventing shortcut learning through gradient isolation. This enables safe transformation of historical trajectories into robust planning priors across diverse driving environments.

	\item We propose the History-Future Trajectory Dictionary Network (HFTDN), which constructs a cross-dataset trajectory dictionary and employs historical trajectory similarity-based retrieval to identify relevant future trajectories. The Universal Dataset Trajectory Guide Module transforms retrieved trajectories into dataset-agnostic planning guidance, enabling the planner to leverage diverse driving experiences and maneuvers from multiple datasets.
	
	\item We propose the Sparse-to-Dense (S2D) training paradigm that adaptively masks priors during training to prevent over-reliance, while fully exploiting them during inference for optimal performance.
	
	\item Extensive evaluation on nuPlan validates our framework's effectiveness, with UniPlanner achieving significant improvements on both Test14-random (NR-CLS: 87.25, +4.14\%; R-CLS: 85.25, +3.96\%) and Test14-hard (NR-CLS: 71.38, +3.63\%; R-CLS: 70.99, +2.25\%) benchmarks. Comprehensive ablation studies validate the essential contribution of each component, confirming that performance gains directly stem from our multi-dataset integration approach. These results establish cross-dataset knowledge aggregation as a promising paradigm for advancing motion planning capabilities.
	
\end{itemize}

As the first motion planning framework to achieve multi-dataset integration for autonomous vehicle decision-making, UniPlanner provides key insights into cross-dataset learning challenges and establishes a scalable paradigm for training robust systems.
The paper is organized as follows: Section \ref{sec:related work} reviews related work, Section \ref{sec:method} details the proposed framework, Section \ref{sec:experiment} presents experimental evaluation, and Section \ref{sec:conclusion} conclusion.

\section{RELATED WORKS}
\label{sec:related work}
\subsection{Motion Planning for Autonomous Vehicles}
Recent years have witnessed significant advances in learning-based motion planning for autonomous vehicles.
Motion planning in autonomous vehicles must handle discrete-continuous action spaces, training instability, and safety-critical uncertainties while generating executable trajectories \cite{ref1}. To address the discrete-continuous challenge, Ni et al. \cite{ref1} coupled DDQN \cite{ref55} for lane-changing with TD3 \cite{ref57} for car-following, while Chen et al. \cite{ref2} stabilized this hybrid training through imitation learning initialization. Building on these solutions, Yang et al. \cite{ref53} further integrated uncertainty quantification with fallback MPC policies to ensure safety under perception errors.

However, driving is inherently a group behavior where predicting surrounding vehicles' motions is crucial for safe planning, leading to various interaction-aware approaches. Huang et al. \cite{ref4} proposed differentiable joint prediction-planning to capture these interactions. Game-theoretic approaches \cite{ref56} offered alternative principled frameworks. For instance, GameFormer \cite{ref5} modeled driver sophistication through hierarchical reasoning, while Ma et al. \cite{ref3} achieved real-time performance by decomposing complex scenarios into pairwise leader-follower games.

To effectively model these complex interactions, most methods rely heavily on historical trajectories as crucial prior information for understanding driving patterns and predicting future behaviors \cite{ref4}\cite{ref5}. These temporal priors enable models to capture motion patterns and anticipate agent intentions, significantly improving open-loop performance. However, recent studies have uncovered a fundamental problem with this dependency: the same historical information that enhances prediction accuracy causes catastrophic shortcut learning during closed-loop deployment \cite{ref22}\cite{ref24}. Models learn to exploit trajectory patterns rather than understanding actual driving logic. To address this problem, Guo et al. \cite{ref27} completely abandoned historical inputs and redesigned the coordinate system, while Cheng et al. \cite{ref28} employed state dropout to break the dependency. Unfortunately, discarding this prior information means losing valuable driving knowledge that experts naturally leverage for robust decision-making.

Beyond these technical challenges, all existing learning-based methods share a fundamental limitation: training on single datasets restricts their adaptability across diverse traffic environments. UniPlanner overcomes both the historical information paradox and the generalization limitation through a novel combination of multi-dataset integration with gradient-free learning, which uniquely enables safe extraction of transferable expert knowledge while preventing shortcut dependencies that plague gradient-based approaches.

\begin{figure*}[!t]
	\centering
	\includegraphics[width=6 in]{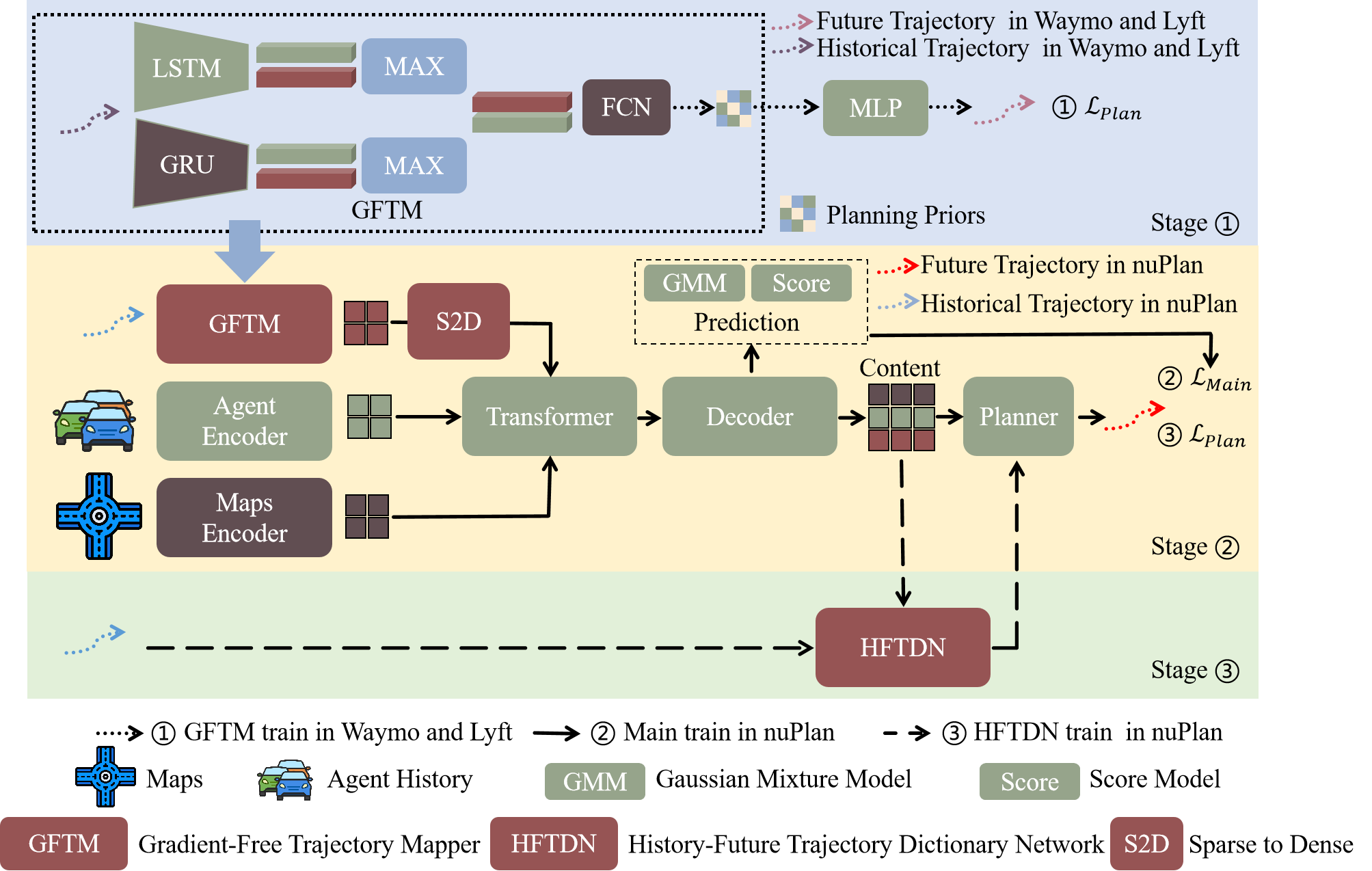}
	\caption{Overview of UniPlanner architecture with three core modules (GFTM, HFTDN, and S2D) and the sequential training pipeline. \scalebox{1.1}{\ding{192}}, \scalebox{1.1}{\ding{193}}, and \scalebox{1.1}{\ding{194}} denote GFTM training, main network training, and HFTDN training phases, respectively}
	
	\label{fig:3}
\end{figure*}

\subsection{Multi-Dataset Learning for Autonomous Vehicles}
Multi-dataset learning leverages diverse data sources to enhance generalization in robotics and autonomous driving \cite{ref10}, addressing the fundamental challenge of data scarcity through transfer learning and joint training strategies.

Initial successes emerged in mobile robotics, where Shah et al. \cite{ref10}\cite{ref11} pioneered multi-dataset training with waypoint normalization and transformers \cite{ref12}, while Sridhar et al. \cite{ref13} achieved robust navigation using diffusion models \cite{ref14}. However, these unmanned platforms benefit from relatively homogeneous sensor data \cite{ref15}\cite{ref16} and operate in unstructured environments without the stringent safety requirements of autonomous vehicles.

Applying multi-dataset learning to autonomous vehicles presents unique challenges. Vehicle datasets exhibit significant heterogeneity in sensor configurations, coordinate systems, and annotation formats across platforms like Waymo \cite{ref19}, Lyft \cite{ref18},  and nuPlan \cite{ref17}. Consequently, existing vehicle-related work remains limited to narrow applications: Yosinski et al. \cite{ref43} explored feature transfer, Hou et al. \cite{ref42} addressed steering control. Feng et al. \cite{ref25} proposed UniTraj, a unified framework for scalable vehicle trajectory prediction through multi-dataset training. Jiang et al. \cite{ref46} construct a dense interaction dataset by systematically mining interaction events from large-scale naturalistic driving trajectories collected across different datasets, enabling comprehensive analysis of diverse interaction patterns. Yet multi-dataset motion planning, essential for vehicle decision-making, remains unexplored.

By exploiting trajectory distribution invariances and temporal correlations, UniPlanner enables robust multi-dataset integration for motion planning, establishing the first comprehensive framework for cross-dataset learning in autonomous vehicles.

\section{METHOD}
\label{sec:method}
This section introduces UniPlanner, a unified planning framework that realizes cross-dataset integration, transforming diverse driving data into universal planning prior and guidance. UniPlanner adopts an integrated prediction-planning architecture (Figure \ref{fig:3}), directly utilizing GameFormer's \cite{ref5} prediction module for surrounding vehicle trajectory forecasting.
Our core innovations lie in the planning module, which comprises three key components: (1) the Gradient-Free Trajectory Mapper (GFTM) that captures universal history-to-future correlations across datasets, (2) the History-Future Trajectory Dictionary Network (HFTDN) that aggregates history-future pairs from multiple datasets to enable cross-dataset retrieval and guidance, and (3) the Sparse-to-Dense (S2D) training paradigm that balances robustness with prior utilization. These components synergistically transform cross-dataset knowledge into a unified planning framework.

\subsection{Gradient-Free Trajectory Mapper}
Historical trajectories serve as crucial planning information, encoding rich spatiotemporal information about driving behaviors and maneuver patterns. Prior work has identified strong history-future correlations in driving trajectories \cite{ref22,ref24}. Through extensive empirical analysis (Figure \ref{fig:2}), we validate that these correlations universally exist across diverse autonomous driving datasets, demonstrating that past motion states fundamentally shape future planning decisions. However, while these correlations provide invaluable prior for trajectory planning, they will induces shortcut learning that compromises performance capability \cite{ref4}\cite{ref5}.

To resolve this fundamental trade-off, we propose the Gradient-Free Trajectory Mapper (GFTM), a novel module that safely harnesses historical trajectory information while enabling effective cross-dataset learning (Figure \ref{fig:3}). The design of GFTM is motivated by two fundamental discoveries: first, trajectory representations maintain universal consistency across datasets despite sensor heterogeneity; second, the statistical correlations between historical and future trajectories remain invariant across diverse driving datasets. These insights motivate GFTM to capture these transferable temporal correlations across datasets. Through the learned cross-dataset correlations, GFTM effectively maps historical trajectories to dataset-invariant planning priors for enhanced motion planning.

GFTM employs a dual-branch architecture to capture universal history-future correlations. The LSTM~\cite{ref38} and GRU~\cite{ref39} branches independently extract features from historical trajectories, which are then fused to create a comprehensive temporal representation. Subsequently, an MLP network maps this representation to future trajectories (Figure \ref{fig:3}). During pre-training, this architecture is trained on over 3 million trajectory pairs from the Waymo~\cite{ref19} and Lyft~\cite{ref18} datasets using supervised learning, enabling GFTM to learn fundamental temporal correlations between historical and future trajectories.
 During deployment, the feature extractors, with their gradients frozen to prevent backpropagation, are integrated into the main network. This design leverages the learned history-future correlations to transform historical trajectories into effective planning priors without creating harmful dependencies in the main network.

Through its specialized architecture and gradient-free deployment(Figure \ref{fig:3}), GFTM accomplishes: (1) effective multi-dataset knowledge aggregation via trajectory learning, and (2) safe planning prior utilization through gradient isolation.

\begin{figure}[!t]
	\centering
	\includegraphics[width=3 in]{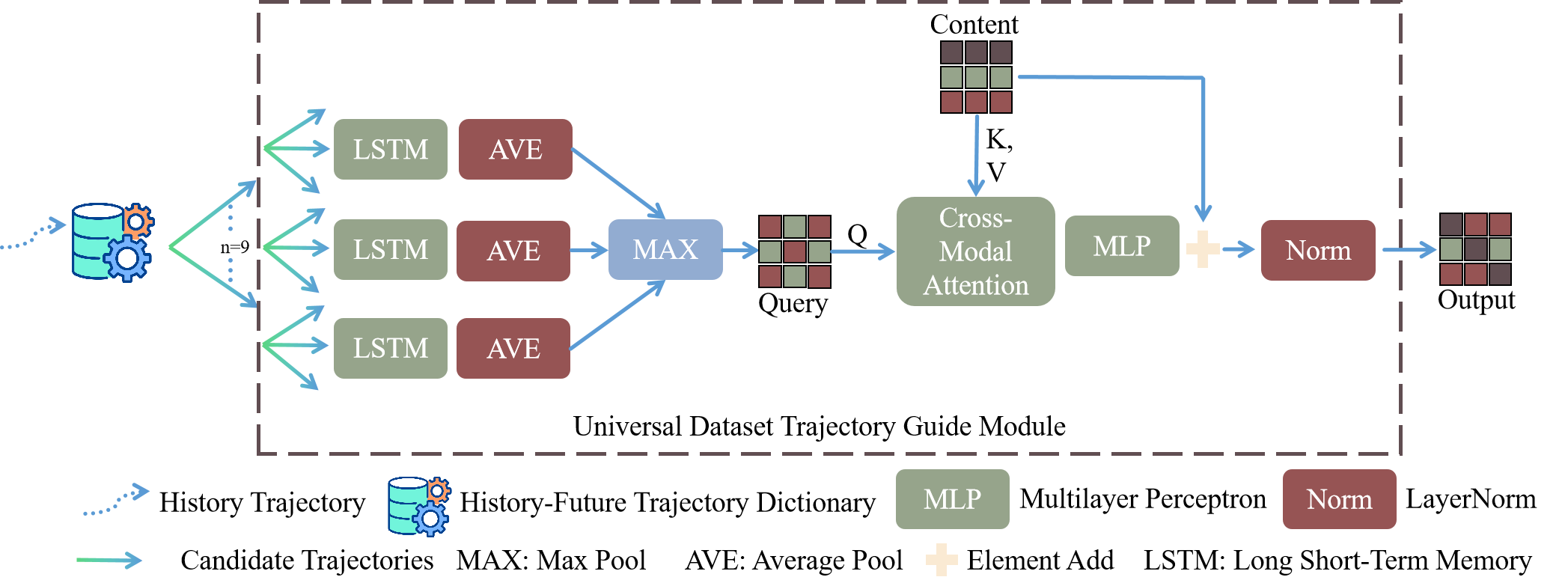}
	\caption{Architecture of the History-Future Trajectory Dictionary Network}
	\label{fig:5}
\end{figure}

\subsection{History-Future Trajectory Dictionary Network}
Building on prior work \cite{ref22}\cite{ref24} and our observation of consistent history-future correlations across datasets (Figure \ref{fig:2}), we propose History-Future Trajectory Dictionary Network (HFTDN), an independent sub-network that performs cross-dataset trajectory integration and retrieval to generate universal planning guidance. HFTDN combines a History-Future Trajectory Dictionary with a Universal Dataset Trajectory Guide Module (Figure \ref{fig:5}). In contrast to GFTM, which learns implicit correlations, HFTDN explicitly retrieves candidate future trajectories through historical trajectory matching, effectively transforming multi-dataset knowledge into actionable planning guidance.

\begin{table*}[ht]
	\centering
	\caption{Distribution of motion types across different datasets and the trajectory dictionary}
	\label{tab:1}
	\begin{tabular}{cccccc}
		\toprule
		\textbf{Dataset} & Stationary & Forward&  Left Turn &    Right Turn	 &  Total\\ 
		\midrule
		
		Waymo\cite{ref19} & 138125	&	325731 &200120	&41365  & 705,341\\
		Lyft\cite{ref18} & 604197	&	1585591 &219403	&84178  & 2,493,369\\
		
		Dictionary & 7 & 1763 & 1385 & 754 & 3909 \\
		
		\bottomrule
	\end{tabular}
\end{table*}

\begin{algorithm}[htbp]
	\caption{History-Future Trajectory Dictionary Construction}
	\label{alg:1}
	\begin{algorithmic}[1]
		\REQUIRE Historical trajectories $\mathscr{H} = \{h_1, ..., h_N\}$, future trajectories $\mathscr{F} = \{f_1, ..., f_N\}$, sampling interval $\Delta t$

		\ENSURE Trajectory dictionary $\mathscr{D}$
		
		\STATE \textbf{// Feature Extraction}
		\FOR{each trajectory pair $(h_i, f_i)$}
		\STATE Extract motion features from $h_i$:
		\STATE \quad $\mathbf{F}_i = [\bar{a}_i, \bar{v}_i, \bar{\kappa}_i, \bar{\omega}_i, \bar{\alpha}_i]$
		\STATE Apply physical constraints:
		\STATE \quad $a_i \leftarrow \text{clip}(a_i, -a_{\max}, a_{\max})$
		\STATE \quad $\kappa_i \leftarrow \text{clip}(\kappa_i, -\kappa_{\max}, \kappa_{\max})$
		\ENDFOR
		
		\STATE \textbf{// Trajectory Binning}
		\STATE Set resolution vector $\mathbf{r} = [2.0, 1.0, 0.02, 0.1, 0.1]$
		
		\FOR{each trajectory pair $(h_i, f_i)$}
		\STATE Compute bin index: $b_i = \lfloor \mathbf{F}_i \oslash \mathbf{r} \rfloor$
		\STATE Assign to bin: $\mathscr{B}_{b_i} \leftarrow \mathscr{B}_{b_i} \cup \{(h_i, f_i)\}$
		\ENDFOR
		
		\STATE \textbf{// Representative Selection via Clustering}
		\FOR{each non-empty bin $\mathscr{B}_k$}
		\STATE Extract historical set: $\mathscr{H}_k = \{h : (h,f) \in \mathscr{B}_k\}$
		\STATE Extract future set: $\mathscr{F}_k = \{f : (h,f) \in \mathscr{B}_k\}$
		\STATE $C_h \leftarrow \text{KMeans}(\mathscr{H}_k, n_{\text{clusters}})$
		\STATE $C_f \leftarrow \text{KMeans}(\mathscr{F}_k, n_{\text{clusters}})$
		\STATE Select representative pairs near cluster centroids
		\STATE $\mathscr{D} \leftarrow \mathscr{D} \cup \{\text{selected pairs}\ AND \  \mathbf{F}_{selected}\}$
		\ENDFOR
		
		\RETURN $\mathscr{D}$
	\end{algorithmic}
\end{algorithm}

\subsubsection{History-Future Trajectory Dictionary Construction }

We construct the trajectory dictionary using over 3 million history-future trajectory pairs extracted from Waymo \cite{ref19} and Lyft \cite{ref18} datasets (Table \ref{tab:1}). Given the substantial redundancy and distributional imbalance inherent in this large-scale collection, we employ bin-based clustering to identify representative samples. Algorithm \ref{alg:1} details our dictionary construction methodology, which ensures balanced coverage of diverse trajectory  while significantly reducing computational complexity.

\begin{figure}[!t]
	\centering
	\includegraphics[width=3 in]{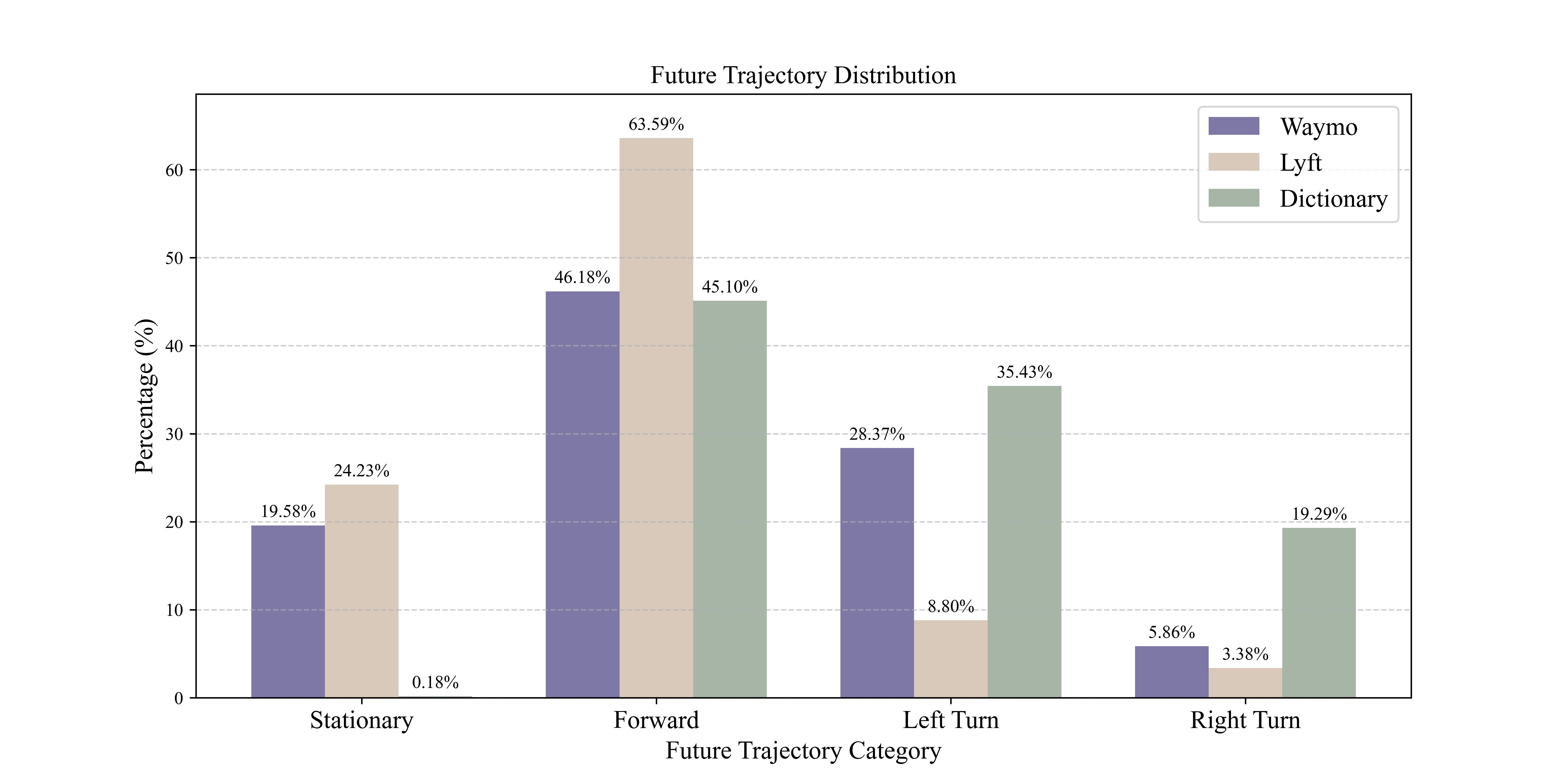}
	\caption{Comparison of motion type distributions between source datasets and dictionary
	}
	\label{fig:6}
\end{figure}

Unlike prior works \cite{ref29}\cite{ref36} that perform binning based on current ego states, we employ trajectory binning using historical trajectory , enabling comprehensive  motion feature extraction and fine-grained behavior categorization.
We extract motion features $\mathbf{F} = [\bar{a}, \bar{v}, \bar{\kappa}, \bar{\omega}, \bar{\alpha}]$ from historical trajectories, comprising average acceleration $\bar{a}$, average velocity $\bar{v}$, average curvature $\bar{\kappa}$, average angular velocity $\bar{\omega}$, and average angular acceleration $\bar{\alpha}$. Trajectories with similar feature vectors are assigned to corresponding bins. Within each bin, we apply K-means clustering \cite{ref40} independently to historical and future trajectories, selecting samples near cluster centroids to eliminate outliers and redundancy.

Through this clustering process, we obtain 3,909 representative trajectory pairs forming the dictionary core (Table \ref{tab:1}). These pairs, augmented with their corresponding feature vectors, constitute the complete trajectory dictionary.
Table \ref{tab:1} and Figure \ref{fig:6} presents the dictionary's trajectory distribution, revealing balanced motion representation: 45.28\% for $\textit{Stationary}$ and $\textit{Forward}$ behaviors combined, and 54.72\% for turning maneuvers. This distribution significantly improves upon the original dataset imbalance, ensuring comprehensive coverage of diverse driving behaviors. The uniform binning strategy applied to velocity and acceleration parameters inherently produces fewer bins for $\textit{Stationary}$ trajectories, resulting in their under representation in the initial binning.

\begin{figure}[!t]
	\centering
	\includegraphics[width=3 in]{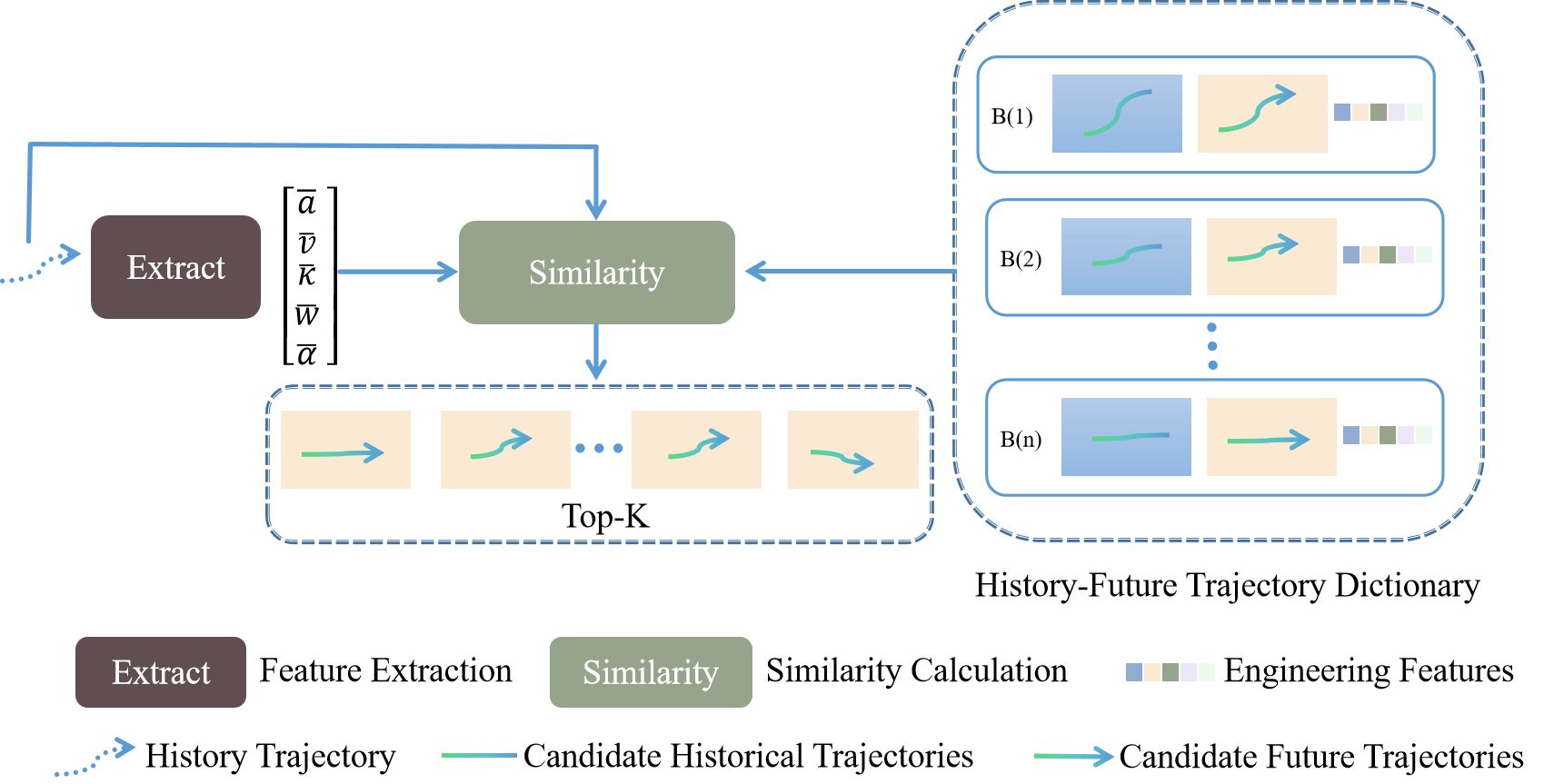}
	\caption{History-based trajectory retrieval process in HFTDN}
	\label{fig:query}
\end{figure}

\subsubsection{History-Based Trajectory Retrieval}
Historical trajectories serve as queries to retrieve the $K$ most similar future trajectories from the dictionary $\mathscr{D}$, as illustrated in Figure \ref{fig:query}. We define a hybrid similarity metric that combines geometric distance and feature similarity:
\begin{equation}
S(h_q, h_i) = \alpha \cdot \text{sim}(h_q, h_i) - (1 - \alpha) \cdot d(h_q, h_i),
\end{equation}
where $h_q$ denotes the query historical trajectory, $h_i$ represents the $i$-th historical trajectory in $\mathscr{D}$, and $\alpha \in [0,1]$ balances the two terms. The normalized cosine similarity $\text{sim}(\cdot,\cdot)$ captures feature-level correspondence, while the normalized L2 distance $d(\cdot,\cdot)$ measures geometric deviation.

The retrieval mechanism identifies the top-$K$ historical matches based on similarity scores $S(h_q, h_i)$ and returns their corresponding future trajectories $\{f_i\}_K$, yielding diverse planning references from the dictionary.

\subsubsection{Universal Dataset Trajectory Guide Module}
Direct incorporation of retrieved future trajectories risks inducing shortcut learning \cite{ref9}. We address this through the Universal Dataset Trajectory Guide Module (UDTGM), which transforms trajectory candidates into universal guidance via grouped encoding, as illustrated in Figure \ref{fig:5}. The UDTGM employs a groupwise processing strategy to process the top-$K$ retrieved candidate trajectories. To manage computational complexity when processing multiple trajectory candidates, we partition the trajectories into $n$ groups, with each group independently encoded by a shared LSTM~\cite{ref38} network. This groupwise processing strategy significantly reduces computational overhead while preserving the ability to capture group-specific motion patterns. Second, intra-group average pooling aggregates temporal features within each group, producing feature representations that preserve local trajectory characteristics. Finally, inter-group max pooling extracts salient features across all groups, generating a unified guidance vector $\mathbf{Q}$.  The unified guidance vector $\mathbf{Q}$ functions as a cross-attention query to inject relevant dictionary knowledge into the current planning context.

To prevent guidance-induced shortcuts, we employ a independent training strategy (Figure \ref{fig:3}). After integrating HFTDN into the planning network, we conduct isolated training with all main network parameters frozen, including GFTM. During this phase, HFTDN receives decoded scene features \cite{ref5} and generates refined representations through the UDTGM. Only HFTDN parameters are updated, ensuring the network learns to utilize guidance without developing dependencies.

This decoupled training paradigm achieves dual objectives: leveraging cross-dataset trajectory knowledge for enhanced planning while maintaining robustness against shortcut learning through controlled gradient flow.

\subsection{Sparse to Dense Paradigm}

\begin{figure}[!t]
	\centering
	\includegraphics[width=3 in]{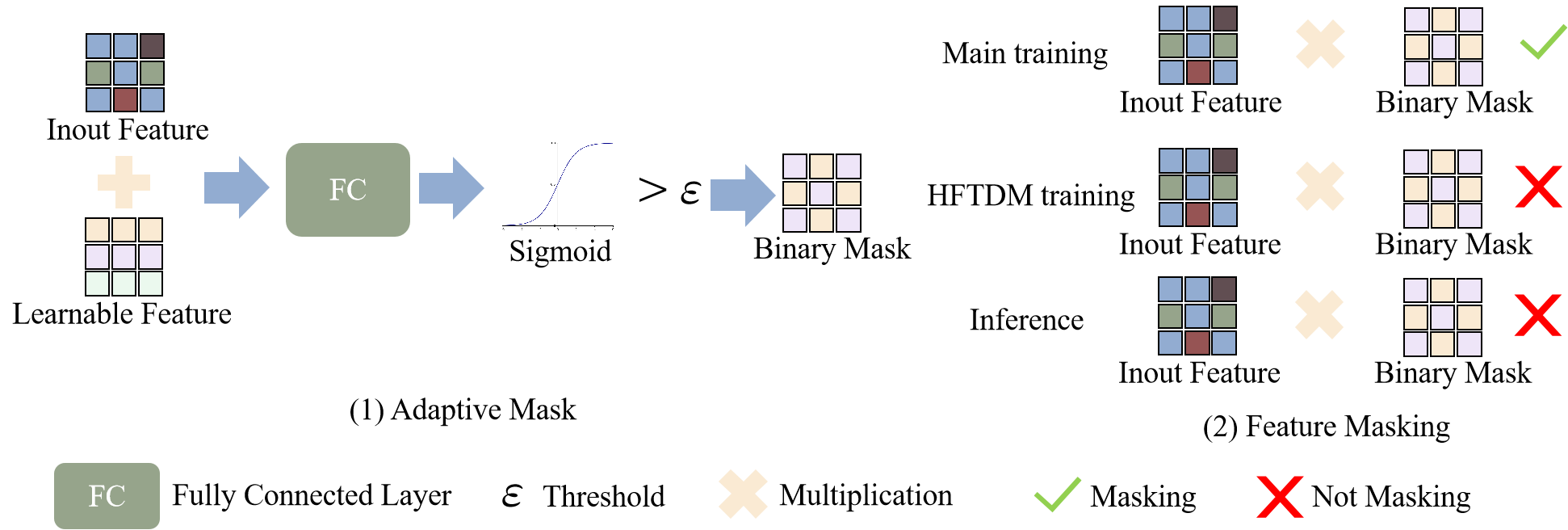}
	\caption{The Sparse-to-Dense training paradigm}
	\label{fig:8}
\end{figure}

While GFTM mitigates direct historical dependencies, residual correlations between planning priors and future trajectories persist, leading to partial shortcut learning that compromises closed-loop performance \cite{ref9}\cite{ref22}\cite{ref24}. To address this limitation, we propose the Sparse-to-Dense (S2D) paradigm, an adaptive suppression strategy that balances shortcut prevention with prior utilization.

S2D employs phase-aware feature masking to optimize prior usage across different operational modes (Figure \ref{fig:8}). The mechanism adapts GFTM's static outputs through learnable parameters $\boldsymbol{\theta}$, enabling dynamic feature modulation despite frozen GFTM weights.
The masking pipeline consists of three steps: (1) learnable parameters augment GFTM outputs via element-wise addition: $\mathbf{F}_a = \mathbf{F}_{GFTM} + \boldsymbol{\theta}$; (2) a sigmoid-activated FC layer converts adaptive features to probabilities $\mathbf{p} \in [0,1]^n$; (3) binary masks are generated through thresholding: $\mathbf{m} = [\mathbf{p} > \epsilon]$ (Figure \ref{fig:8} (1)). The key innovation lies in S2D's operational modes (Figure \ref{fig:8} (2)):
\begin{itemize}
	\item \textbf{Main network training}: Masking activated to suppress shortcuts while learning robust representations.
	\item \textbf{HFTDN training}: Masking deactivated during this phase, as the frozen main network parameters inherently prevent shortcut learning, thereby enabling full utilization of the learned priors.
	\item \textbf{Inference}: Masking deactivated to maximize planning performance through dense prior exploitation.
\end{itemize}

This design ensures that planning priors contribute optimally across all phases: preventing over-reliance during training while enabling comprehensive utilization during deployment, thereby achieving superior planning performance without shortcut-induced degradation.

\subsection{Scene Encoding and Decoding }
The encoding of scene and surrounding agents follows GameFormer \cite{ref5}. The scene encoder processes multi-modal inputs comprising historical state information $S_p \in \mathbb{R}^{N \times T_h \times d_s}$ and vectorized map polylines $M \in \mathbb{R}^{N \times N_m \times N_p \times d_p}$, where $N$, $T_h$, $d_s$, $N_m$, $N_p$, and $d_p$ denote the number of surrounding agents, historical time steps, state attribute dimensions, number of map elements per agent, waypoints per element, and waypoint attribute dimensions, respectively.

Agent encoding employs LSTM networks to process historical trajectories $S_p$, producing tensor $A_p \in \mathbb{R}^{N \times D}$. Map encoding utilizes an MLP to generate features $M_p \in \mathbb{R}^{N \times N_m \times N_p \times D}$ from polylines, which are then aggregated via max-pooling to $M_r \in \mathbb{R}^{N \times N_{mr} \times D}$, where $N_{mr}$ represents the number of aggregated map elements and $D$ denotes the hidden feature dimension.

Agent features are concatenated with corresponding local map features to form agent-wise scene context tensor $C^i = [A_p, M_r^i] \in \mathbb{R}^{(N+N_{mr}) \times D}$ for agent $i$, where all agents' features $A_p$ are included. This is processed through a transformer encoder with $E$ layers. The final scene context encoding $C_s \in \mathbb{R}^{N \times (N+N_{mr}) \times D}$ incorporates all agents for subsequent decoding.

The decoder and prediction modules adhere to the architectural design of GameFormer \cite{ref5}. Given that prediction refinement is beyond the purview of this investigation, we utilize a single-step prediction paradigm to produce the final trajectory output.

\subsection{Hierarchical Loss Functions}
 Our framework employs a three-stage training pipeline to optimize each component independently, as illustrated in Figure \ref{fig:3}.
 
 \subsubsection{Phase I: GFTM Pre-training}
 We train GFTM on external datasets to learn history-to-future mappings using a composite planning loss:
 \begin{equation}
 \mathscr{L}_{\text{Plan}} = \mathscr{L}_{\text{Huber}}(\hat{\mathbf{T}}, \mathbf{T}) +  \mathscr{L}_{\text{Huber}}(\hat{\psi}, \psi),
 \end{equation}
 where $\hat{\mathbf{T}}$ and $\mathbf{T}$ denote predicted and ground-truth trajectories, $\hat{\psi}$ and $\psi$ represent predicted and ground-truth yaw angles.  $\mathscr{L}_{\text{Huber}}$ denotes the Huber loss \cite{ref31}, chosen for its robustness to outliers in trajectory planning.

 \subsubsection{Phase II: Main Network Training}
 The main network jointly optimizes ego-vehicle planning and surrounding agent prediction:
 \begin{equation}
 \mathscr{L}_{\text{Main}} = \mathscr{L}_{\text{Plan}} + \mathscr{L}_{\text{Pred}},
 \end{equation}
 where $\mathscr{L}_{\text{Plan}}$ supervises ego-vehicle trajectory planning following the same formulation as GFTM training. The prediction loss $\mathscr{L}_{\text{Pred}}$ supervises multi-agent trajectory forecasting using negative log-likelihood of Gaussian mixture models, as formulated in GameFormer \cite{ref5}.

 \subsubsection{Phase III: HFTDN Fine-tuning}
 With the main network frozen, we optimize HFTDN using the planning loss $\mathscr{L}_{\text{Plan}}$ to ensure effective guidance generation without inducing shortcuts.

\section{EXPERIMENTS}
\label{sec:experiment}
\subsection{Datasets} 
Our experiment employs three comprehensive autonomous driving datasets: nuPlan \cite{ref17}, Waymo \cite{ref19}, and Lyft \cite{ref18}. 

\subsubsection{Primary Dataset} 
 \textbf{nuPlan Dataset} \cite{ref17} constitutes the first large-scale planning benchmark for autonomous driving, encompassing 1,200 hours of human driving data collected across four geographically diverse cities: Boston, Pittsburgh, Las Vegas, and Singapore. Each city exhibits distinct traffic scene and driving behaviors, providing comprehensive coverage of various driving scenarios.

\subsubsection{Auxiliary Datasets}
 \textbf{Waymo Motion Dataset} \cite{ref19} comprises 103,354 driving scenes with complete object trajectories and 3D map information. The dataset contains over 570 hours of driving data, with each scene spanning 20 seconds sampled at 10 $ Hz $, collected across 1,750 km of roadways in six major U.S. cities.

 \textbf{Lyft Level 5 Dataset} \cite{ref18} provides over 1,000 hours of autonomous driving data captured by a fleet of 20 self-driving vehicles. It includes 15,242 semantically annotated map elements and high-definition aerial imagery, specifically designed to advance motion prediction, planning, and simulation research.

\subsubsection{Data Preprocessing} 
nuPlan \cite{ref17} serves as our primary dataset for training both the main network and the  HFTDN, while Waymo \cite{ref19} and Lyft \cite{ref18} are utilized exclusively for GFTM training and trajectory dictionary construction. We adopt the preprocessed nuPlan dataset from Chen et al. \cite{ref30}, which preserves complete perception information. For the Lyft~\cite{ref18} and Waymo~\cite{ref19} datasets, we adopt the same preprocessing procedures~\cite{ref5}, extracting only trajectory data while excluding  perception data. Given the exceptionally large volume of trajectory data in the Lyft \cite{ref18}, we implement a random sampling strategy to retain a representative subset, ensuring computational feasibility while maintaining dataset diversity. All trajectory data are standardized to 10 $ Hz $ sampling rate, comprising 2-second historical trajectories with state vectors $(x, y, \psi, v, \omega)$ and 8-second future trajectories with state vectors $(x, y, \psi)$, where $ (x ,  y) $ is the position, $\psi$ denotes heading angle, $v$ represents velocity, and $\omega$ indicates angular velocity.

\subsection{Implementation Details}

Our training methodology consists of three sequential phases, as illustrated in Figure \ref{fig:3}. 

\textbf{Phase I: GFTM Pre-training.} The GFTM is initially trained on the combined Waymo \cite{ref19} and Lyft \cite{ref18} datasets for 50 epochs using a batch size of 1,000 and a learning rate of $2 \times 10^{-3}$. Upon completion, we extract the GFTM  (excluding the MLP head) and integrate them into the main network with frozen parameters.

\textbf{Phase II: Main Network Training.} The main network undergoes training on the nuPlan dataset \cite{ref17} for 50 epochs with a batch size of 196 and a learning rate of $2 \times 10^{-4}$. During main network training, S2D selectively suppresses GFTM outputs through adaptive masking, forcing the network to learn robust features rather than developing shortcut dependencies.

\textbf{Phase III: HFTDN Training.} In this phase, all main network parameters are frozen and the adaptive mask is removed. The HFTDN is then integrated with the main network, and the complete system is trained on nuPlan \cite{ref17} for 50 epochs using a batch size of 512 and a learning rate of $1 \times 10^{-4}$. The Universal Dataset Trajectory Guide Module transforms dictionary-retrieved trajectories into cross-dataset planning guidance, enhancing decision-making performance.

All experiments are conducted on an NVIDIA RTX 4090 GPU. Table \ref{tab:param} summarizes the key hyperparameters employed in our implementation.
\begin{table}[ht]
	\centering
	\caption{HYPERPARAMETERS OF THE MODEL AND TRAINING PROCESS}
	\label{tab:param}
	\begin{tabular}{ccc}
		\toprule
		Type & Parameters & Value \\ 
		\midrule
		& Batch size & 1000\\
		GETM training	& Learning rate & 0.002\\
		& Epochs & 50\\
		\midrule
		&Batch size & 196\\
		Main training	&Learning rate & 0.0002\\
		& Epochs & 50\\
		\midrule
		&Batch size & 512\\
		HFTDN training	&Learning rate & 0.0001\\
		&Epochs & 50\\
		\midrule
		&Similarity weight $\alpha$ &  0.3\\
		&	Thresholding $ \epsilon $ & 0.7\\
		&	Maximum curvature $\kappa $ & 0.5\\
		&	Maximum acceleration $a $ & 5\\
		&	 Time interval $dt $ & 0.1\\
		Model parameters&     Clusters $ n_{\text{clusters}} $    & 2 \\
		
		&	$ Top-K $  & 9\\
		&	Group $ n $ & 3\\
		&	History steps  & 20\\
		&	Future steps  & 80\\
		
		\bottomrule

	\end{tabular}
\end{table}

\subsection{Evaluation} 
This section presents a comprehensive comparative analysis of UniPlanner against state-of-the-art planning methods to demonstrate the effectiveness of our approach.

\subsubsection{Baseline Methods} 
We evaluate UniPlanner against ten representative autonomous driving planners:

(1) \textbf{RasterModel} \cite{ref17}: A convolutional neural network based planner that processes rasterized representations of the driving scene.

(2) \textbf{UrbanDriver} \cite{ref32}: A transformer-based architecture that leverages polyline encoders for structured scene representation.

(3) \textbf{IDM} (Intelligent Driver Model) \cite{ref33} : A commonly used vehicle following model and trajectory planning method in the fields of autonomous driving and traffic simulation.

(4) \textbf{PDM} \cite{ref7}: The winning solution of the 2023 nuPlan Planning Challenge. We evaluate PDM-Closed, which integrates rule-based IDM \cite{ref33} for enhanced safety.

(5) \textbf{GC-PGP} \cite{ref34}: A goal-conditioned planner that performs lane graph traversals for structured navigation.

(6) \textbf{GameFormer} \cite{ref5}: An interactive prediction and planning framework that incorporates level-k game theory for multi-agent reasoning.

(7) \textbf{PlanTF} \cite{ref9}: An efficient transformer-based architecture designed for imitation learning in autonomous driving.

(8) \textbf{DTPP} \cite{ref8}: A differentiable trajectory planning framework that combines trajectory sampling with deep learning optimization.

(9) \textbf{InstructDriver} \cite{ref35}: A novel approach that integrates large language models (LLMs) for interpretable motion planning.

(10) \textbf{PPP} \cite{ref48}:  A planner that improves trajectory robustness by fusing multimodal path predictions with environmental features and using multi-stage trajectory evaluation.

\subsubsection{Evaluation Metrics}
We adopt the official nuPlan \cite{ref17} evaluation framework, utilizing two primary metrics: Non-Reactive Closed-Loop Score (NR-CLS) and Reactive Closed-Loop Score (R-CLS). While both metrics share identical trajectory evaluation methodologies, they differ in simulation complexity. NR-CLS evaluates planning performance with static background agents, whereas R-CLS introduces dynamic traffic agents that respond to ego-vehicle actions \cite{ref33}, enabling comprehensive assessment of interactive planning capabilities in realistic traffic scenarios.

\begin{table}[ht]
	\centering
	\caption{Comparison with classic in Test14-random benchmark}
	\label{tab:2}
	\begin{tabular}{ccc}
		\toprule
		\textbf{Planners} & NR-CLS & R-CLS  \\ 
		
		\midrule
		IDM\cite{ref33} &   86.48 &  80.59 \\
		PDM-Closed\cite{ref7}        &    \textbf{90.05} &  \textbf{91.64} \\

		RasterModel\cite{ref17} &  69.66 &67.54\\
		UrbanDriver\cite{ref32} &   63.27 & 61.02\\
		GC-PGP\cite{ref34} &   55.99 &  51.39 \\
		
		GameFormer\cite{ref5} &   80.33 &  81.75 \\
		PlanTF\cite{ref9} &   86.40 &  80.59 \\
		InstructDriver\cite{ref35}&   70.31 &  66.96 \\
		\midrule
		GameFormer\cite{ref5}+ours & 81.9 (+1.57\%) &81.40 (-0.35\%)\\
		UniPlanner (ours)  &   \underline{87.25} &  \underline{85.25} \\           
		
		\bottomrule
	\end{tabular}
\end{table}

\begin{table}[ht]
	\centering
	\caption{Comparison with classic planner in Test14-hard benchmark}
	\label{tab:3}
	\begin{tabular}{ccc}
		\toprule
		\textbf{Planners} & NR-CLS & R-CLS  \\ 
		
		\midrule
		IDM\cite{ref33} &    56.16 &   62.26 \\
		PDM-Closed\cite{ref7}        &     65.07 &  \textbf{75.18} \\

		RasterModel\cite{ref17} &  49.47 &52.16\\
		UrbanDriver\cite{ref32} &   51.54 & 49.07\\
		GC-PGP\cite{ref34} &   43.22 &   39.63 \\
		
		GameFormer\cite{ref5} &   64.68 &  65.70  \\
		PlanTF\cite{ref9} &   \textbf{72.68} &  61.70 \\
		DTPP\cite{ref8}&   59.44 &  62.94 \\
		InstructDriver\cite{ref35}&   57.37 &  52.95 \\
		PPP \cite{ref48}& 65.17&69.27\\
			\midrule
		GameFormer\cite{ref5}+ours & 65.84 (+1.16\%)& 67.85 (+2.15\%)\\
		
		UniPlanner (ours)  &   \underline{71.38} &  \underline{70.99} \\           
		
		\bottomrule
	\end{tabular}
\end{table}

\subsubsection{Benchmark }
Following the evaluation protocol established in \cite{ref9}, we assess UniPlanner on two complementary nuPlan benchmarks: (1) \textbf{Test14-random}, comprising representative driving scenarios encountered in typical urban environments, and (2) \textbf{Test14-hard}, specifically curated to evaluate performance on challenging long-tail scenarios that test the robustness of planning algorithms.

\subsubsection{Quantitative Results}

\begin{figure*}[!t]
	\centering
	\includegraphics[width=6in]{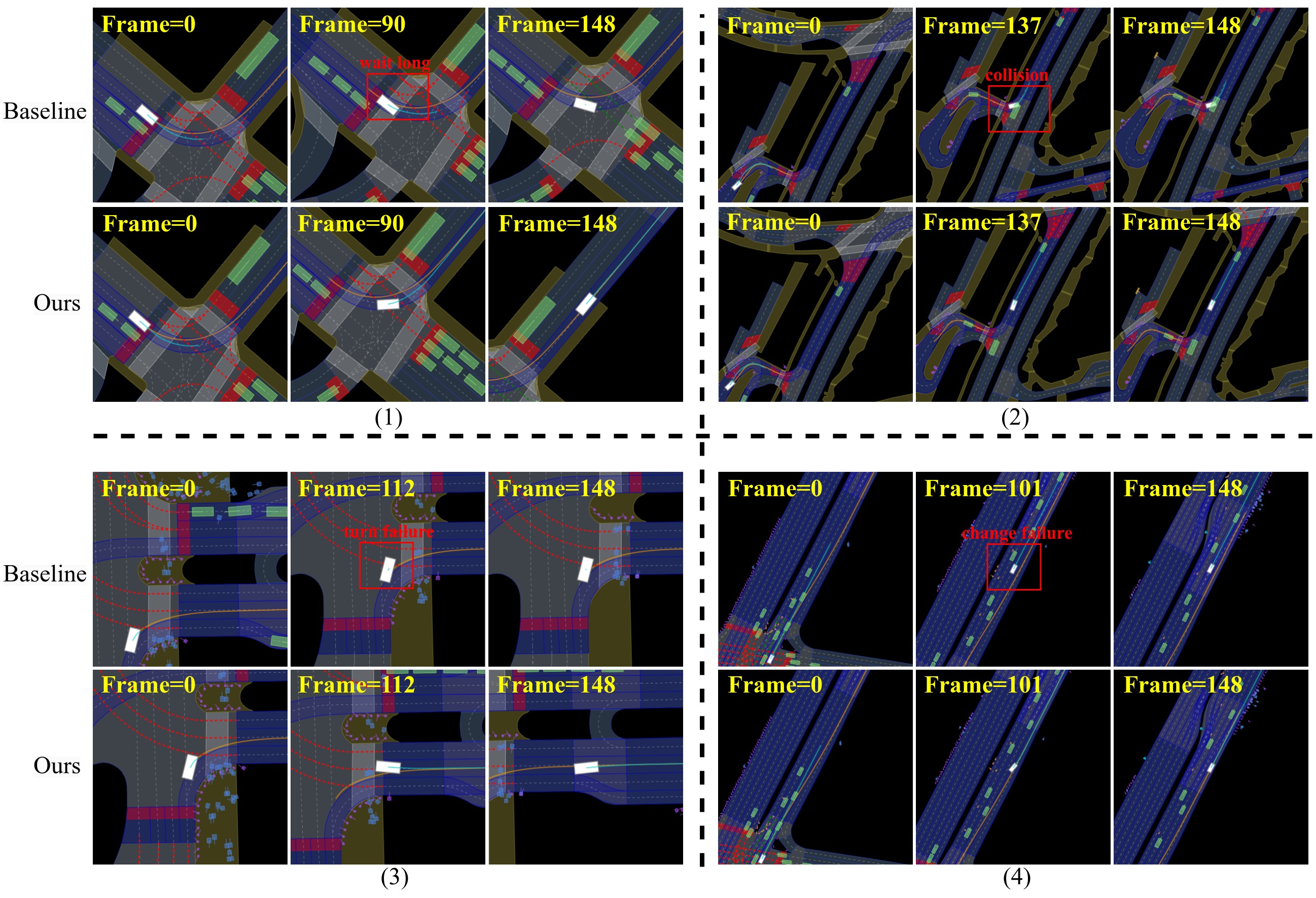}
	\caption{Visualization comparison of planning results on Test14-random benchmark}
	\label{fig:9}
\end{figure*}

Tables \ref{tab:2} and \ref{tab:3} present comprehensive quantitative comparisons between UniPlanner and baseline methods. To ensure fair comparison, performance metrics for competing approaches are sourced from PlanTF \cite{ref9} and their respective original publications. Notably, we retrained Gameformer\cite{ref5} on the same dataset\cite{ref30} to maintain consistency in our evaluation. UniPlanner demonstrates strong performance, achieving the second-highest scores on both R-CLS and NR-CLS metrics across the Test14-random and Test14-hard benchmarks.

To assess the framework compatibility of UniPlanner, we performed integration experiments with GameFormer~\cite{ref5}, a classical planner model. Our results confirm seamless integration while demonstrating consistent performance gains: NR-CLS improves by 1.57\% on Test14-random and by 1.16\% on Test14-hard, with R-CLS showing an even larger 2.15\% improvement on Test14-hard (Tables~\ref{tab:2} and \ref{tab:3}). Although UniPlanner does not achieve state-of-the-art absolute performance, our primary contribution is establishing a novel cross-dataset knowledge transfer paradigm that effectively leverages planning priors from multiple datasets to enhance decision-making in autonomous driving systems.

\subsubsection{Qualitative evaluation}
We demonstrate UniPlanner's superior planning capabilities through qualitative analysis of four representative scenarios from the Test14-random benchmark. Figure \ref{fig:9} presents comparative visualizations between UniPlanner and the baseline planner across these challenging driving situations.

\textbf{Scenario 1: High-Curvature Turn.} This scenario demands significant lateral acceleration for successful navigation. The baseline planner exhibits context misinterpretation, resulting in vehicle immobilization and failure to execute the turn. In contrast, UniPlanner accurately assesses the driving context, initiates timely lateral acceleration, and successfully completes the turning maneuver.

\textbf{Scenario 2: Right Turn.} In dense traffic during a right turn, the baseline planner generates unsafe trajectories resulting in vehicle collisions. UniPlanner, however, demonstrates superior spatial reasoning by executing a collision-free turn while maintaining safe distances from surrounding traffic.

\textbf{Scenario 3: Pedestrian Crossing.} This scenario involves dynamic pedestrian interaction at an intersection. While the baseline planner correctly gives way to crossing pedestrians, it fails to recognize the subsequent turning opportunity, resulting in indefinite immobilization. UniPlanner demonstrates advanced temporal reasoning by appropriately giving way to pedestrians and subsequently identifying a safe gap to complete the turning maneuver.

\textbf{Scenario 4: Lane Change.} The final scenario requires a mandatory lane change for route completion. The baseline planner fails to initiate the lane change, maintaining its current lane and resulting in route deviation. UniPlanner successfully recognizes the lane change requirement, plans an appropriate trajectory, and executes the maneuver within the available traffic gap.

These qualitative results validate that UniPlanner successfully leverages universal trajectory knowledge from multiple datasets to guide planning decisions, demonstrating that cross-dataset history-future correlations can be effectively transformed into actionable planning priors for enhanced performance across diverse driving scenarios.

\begin{figure*}[!t]
	\centering
	\includegraphics[width=6 in]{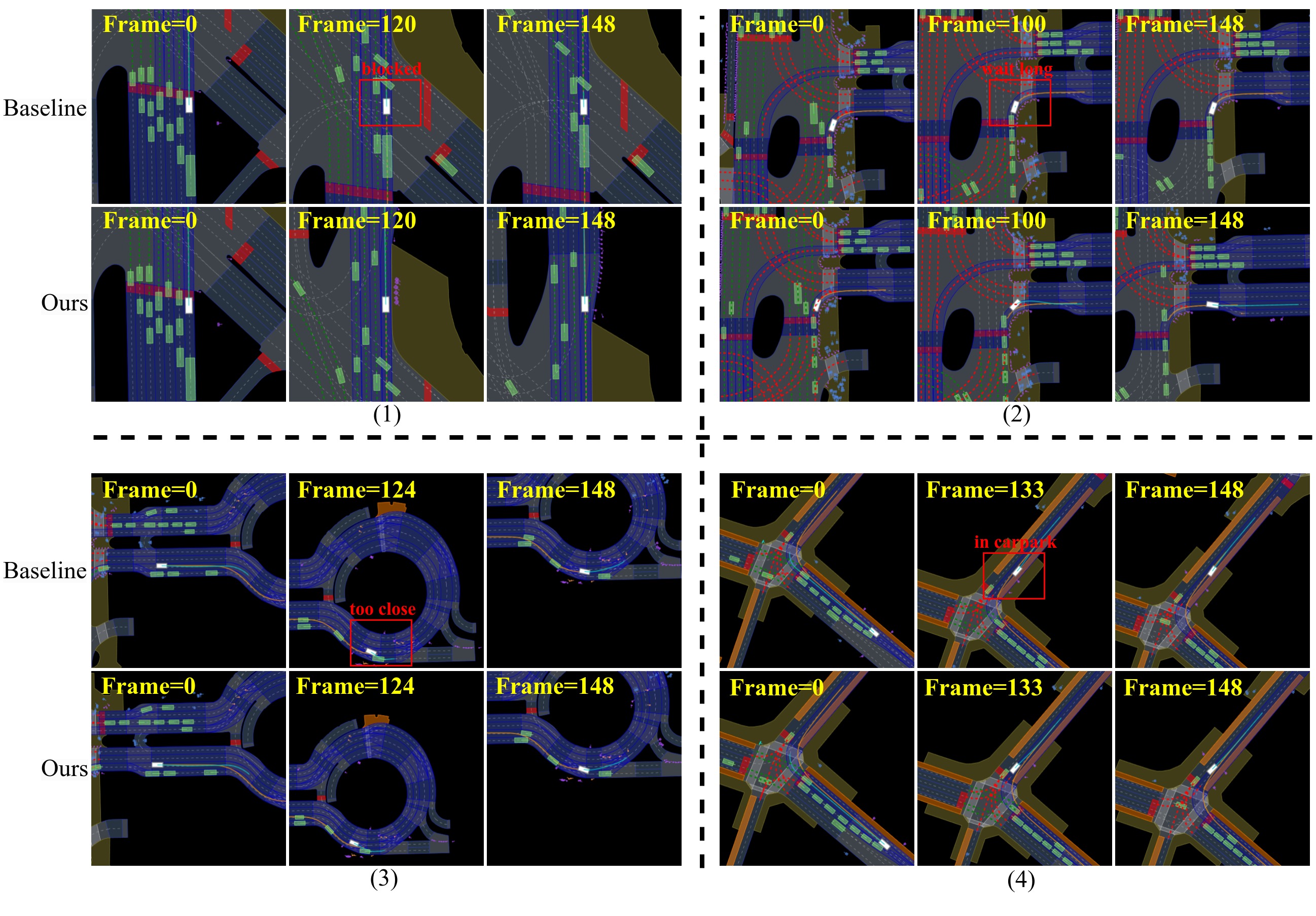}
	\caption{Visualization comparison of planning results on Test14-hard benchmark
	}
	\label{fig:10}
\end{figure*}

To further validate UniPlanner's robustness, we analyze four challenging long-tail scenarios from the Test14-hard benchmark, with comparative results presented in Figure \ref{fig:10}.

\textbf{Scenario 1: High-Speed Accident Avoidance.} This scenario involves high-speed navigation approaching a developing accident scene. The baseline planner exhibits excessive caution, decelerating prematurely and subsequently becoming trapped between surrounding vehicles when the accident materializes. UniPlanner demonstrates superior situational awareness by maintaining appropriate speed and successfully traversing the area before the accident develops, thereby reaching the destination without incident.

\textbf{Scenario 2: Pedestrian-Regulated Turn.} At an intersection with active pedestrian crossing, the baseline planner correctly yields to pedestrians but suffers from decision paralysis post-clearance, resulting in indefinite immobilization. UniPlanner exhibits balanced decision-making by appropriately yielding during pedestrian presence and subsequently executing a decisive turning maneuver upon safe clearance.

\textbf{Scenario 3: U-Turn.} Both planners successfully complete the U-turn maneuver; however, the baseline trajectory passes dangerously close to surrounding vehicles, creating collision risks. UniPlanner generates a safer trajectory that maintains adequate safety margins throughout the maneuver, demonstrating superior spatial planning under geometric constraints.

\textbf{Scenario 4: Right Turn with Spatial Constraints.} During a constrained right turn, the baseline planner generates a trajectory that encroaches upon the adjacent parking area, violating spatial boundaries and creating potential safety hazards. UniPlanner adheres strictly to the drivable area constraints, executing the turn while respecting all spatial boundaries.

The qualitative evaluations across both Test14-random and Test14-hard benchmarks validate UniPlanner's core innovation: successfully integrating trajectory knowledge from multiple datasets to enhance planning performance. These results demonstrate that our dual-module architecture, combining GFTM's gradient-free universal priors with HFTDN's cross-dataset trajectory guidance, enables robust decision-making even in challenging long-tail scenarios. By leveraging universal history-future correlations across diverse datasets, UniPlanner establishing a new paradigm for motion planning in autonomous driving.

\subsection{Ablation Study}
We conduct ablation studies on Test14-random (Table \ref{tab:4}) and Test14-hard (Table \ref{tab:5}) to analyze the contribution of each component.

\subsubsection{Ablation Study in Test14-random }
Starting from the baseline (NR-CLS: 83.11, R-CLS: 81.29), individual components yield distinct improvements. GFTM provides the largest gains (NR-CLS: 84.42, +1.31\%; R-CLS: 83.40, +2.11\%), demonstrating the effectiveness of cross-dataset planning priors. S2D achieves moderate improvement (NR-CLS: 84.19, +1.08\%; R-CLS: 81.74, +0.45\%) through adaptive feature suppression. HFTDN shows smaller but consistent gains (NR-CLS: 83.48, +0.37\%; R-CLS: 82.35, +1.06\%) via trajectory retrieval guidance.

Pairwise combinations reveal synergistic effects. HFTDN+GFTM achieves NR-CLS: 85.41 (+2.30\%) and R-CLS: 83.60 (+2.31\%), while GFTM+S2D reaches NR-CLS: 85.36 (+2.25\%) and R-CLS: 83.53 (+2.24\%). However, HFTDN+S2D shows performance degradation (NR-CLS: 81.20, R-CLS: 81.05), indicating potential interference between retrieval-based guidance and feature suppression. This occurs because the baseline learns historical priors after adaptive masking, resulting in incomplete temporal learning. The removal of these masks during HFTDN training exposes the inadequately learned priors, which then provide erroneous guidance instead of useful constraints.
 The complete framework achieves optimal performance: NR-CLS: 87.25 (+4.14\%) and R-CLS: 85.25 (+3.96\%), validating the complementary nature of all three components.

\begin{table}[htbp]
	\centering
	\caption{Ablation study in Test14-random benchmark }
	\label{tab:4}
	\begin{tabular}{ccc|cc}
		\hline
		HFTDN & GFTM & S2D & NR-CLS & R-CLS \\
		\hline
		&      &    & 83.11   &  81.29 \\
		$\checkmark$ &      &   &83.48   &    82.35   \\
		&   $\checkmark$   &   &84.42   &    83.40   \\
		&      &  $\checkmark$ &84.19   &    81.74   \\
		$\checkmark$  &  $\checkmark$    &   &\underline{85.41}   &    \underline{83.60}   \\
		$\checkmark$   &      &  $\checkmark$ &81.20   &   81.05\\
		&     $\checkmark$  &  $\checkmark$ &85.36   &    83.53\\
		
		$\checkmark$  &   $\checkmark$   &  $\checkmark$ &\textbf{87.25}   &   \textbf{85.25}\\
		
		\hline
	\end{tabular}
\end{table}

\begin{table}[htbp]
	\centering
	\caption{Ablation study in Test14-hard benchmark  }
	\label{tab:5}
	\begin{tabular}{ccc|cc}
		\hline
		HFTDN & GFTM & S2D & NR-CLS & R-CLS \\
		\hline
		&      &    & 67.75   &  68.74 \\
		$\checkmark$ &      &   &68.21   &    69.49   \\
		&   $\checkmark$   &   &68.15   &    69.48   \\
		&      &  $\checkmark$ &67.54   &    69.79   \\
		$\checkmark$  &  $\checkmark$    &   &68.01   &    69.87   \\
		$\checkmark$   &      &  $\checkmark$ &66.42   &   69.02\\
		&   $\checkmark$   &  $\checkmark$ &\underline{69.44 }  &    \textbf{71.55}\\
		
		$\checkmark$  &   $\checkmark$   &  $\checkmark$ &\textbf{71.38}   &   \underline{70.99}\\
		
		\hline
	\end{tabular}
\end{table}

\subsubsection{Ablation Study in Test14-hard }
The Test14-hard benchmark, with lower baseline performance (NR-CLS: 67.75, R-CLS: 68.74), reveals different component behaviors under challenging conditions. Individual components show varied impacts: HFTDN and GFTM provide modest gains (HFTDN: NR-CLS 68.21, +0.46\%; GFTM: NR-CLS 68.15, +0.40\%), while S2D exhibits mixed results with decreased NR-CLS (67.54, -0.21\%) but improved R-CLS (69.79, +1.05\%).

Pairwise combinations demonstrate distinct patterns. GFTM+S2D achieves the strongest pairwise performance (NR-CLS: 69.44, +1.69\%; R-CLS: 71.55, +2.81\%), suggesting S2D's feature suppression particularly benefits GFTM priors in complex scenarios. HFTDN+GFTM shows moderate gains (NR-CLS: 68.01, +0.26\%; R-CLS: 69.87, +1.13\%).
The complete framework achieves optimal performance (NR-CLS: 71.38, +3.63\%; R-CLS: 70.99, +2.25\%), with gains exceeding the sum of individual components. 

Our ablation study design ensures clear attribution of performance gains through parameter isolation. HFTDN, trained with frozen main network parameters, demonstrates the effectiveness of cross-dataset feature guidance, while GFTM's frozen integration confirms that its contribution stems from transforming historical trajectories into planning priors. Notably, the superlinear improvement observed when combining these modules validates UniPlanner's synergistic architecture: GFTM provides planning priors, HFTDN delivers trajectory guidance, and S2D prevents shortcut learning, with each component addressing complementary aspects of the planning challenge.

\section{Conclusion}
\label{sec:conclusion}
Through systematic analysis of trajectory distributions and temporal correlations across diverse datasets, we identify key invariances that enable cross-dataset knowledge transfer. Based on these findings, we present UniPlanner, the first planning framework that leverages historical trajectories as a universal medium for multi-dataset integration in motion planning. UniPlanner achieves universal planning through three synergistic components: (1) the Gradient-Free Trajectory Mapper (GFTM), which captures universal history-to-future correlations across datasets and converts historical information into robust planning priors while preventing shortcut learning through gradient isolation; (2) the History-Future Trajectory Dictionary Network (HFTDN), which aggregates history-future trajectory pairs from multiple datasets into a universal dictionary and transforms retrieved trajectories into cross-dataset planning guidance; and (3) the Sparse-to-Dense (S2D) paradigm, which bridges sparse training signals and dense deployment priors to ensure both learning robustness and inference effectiveness.
Experimental results on nuPlan benchmarks demonstrate the effectiveness of our universal approach, achieving 4.14\% and 3.63\% improvements in NR-CLS on Test14-random and Test14-hard, respectively. These results validate that universal trajectory knowledge aggregated from diverse datasets significantly enhances planning robustness. UniPlanner establishes a new paradigm for multi-dataset knowledge integration in motion planning, demonstrating the feasibility and effectiveness of cross-dataset learning while providing a base framework for future research in large-scale dataset aggregation.

\section*{Acknowledgments}
This work was supported in part by the National Natural Science Foundation of China under Grants 62171294, 62101344 and 62372302; in part by the Natural Science Foundation of Guangdong Province, China under Grant 2022A1515010159; in part by the Key project of Shenzhen Science and Technology Plan under Grants 20220810180617001 and JCYJ20200109105832261; and in part by the Tencent ``Rhinoceros Birds'' - Scientific Research Foundation for Young Teachers of Shenzhen University, China.

\bibliographystyle{IEEEtran}
\bibliography{cas-refs}

\section{Biography Section}

 \begin{IEEEbiography}[{\includegraphics[width=1in,height=1.25in,clip,keepaspectratio]{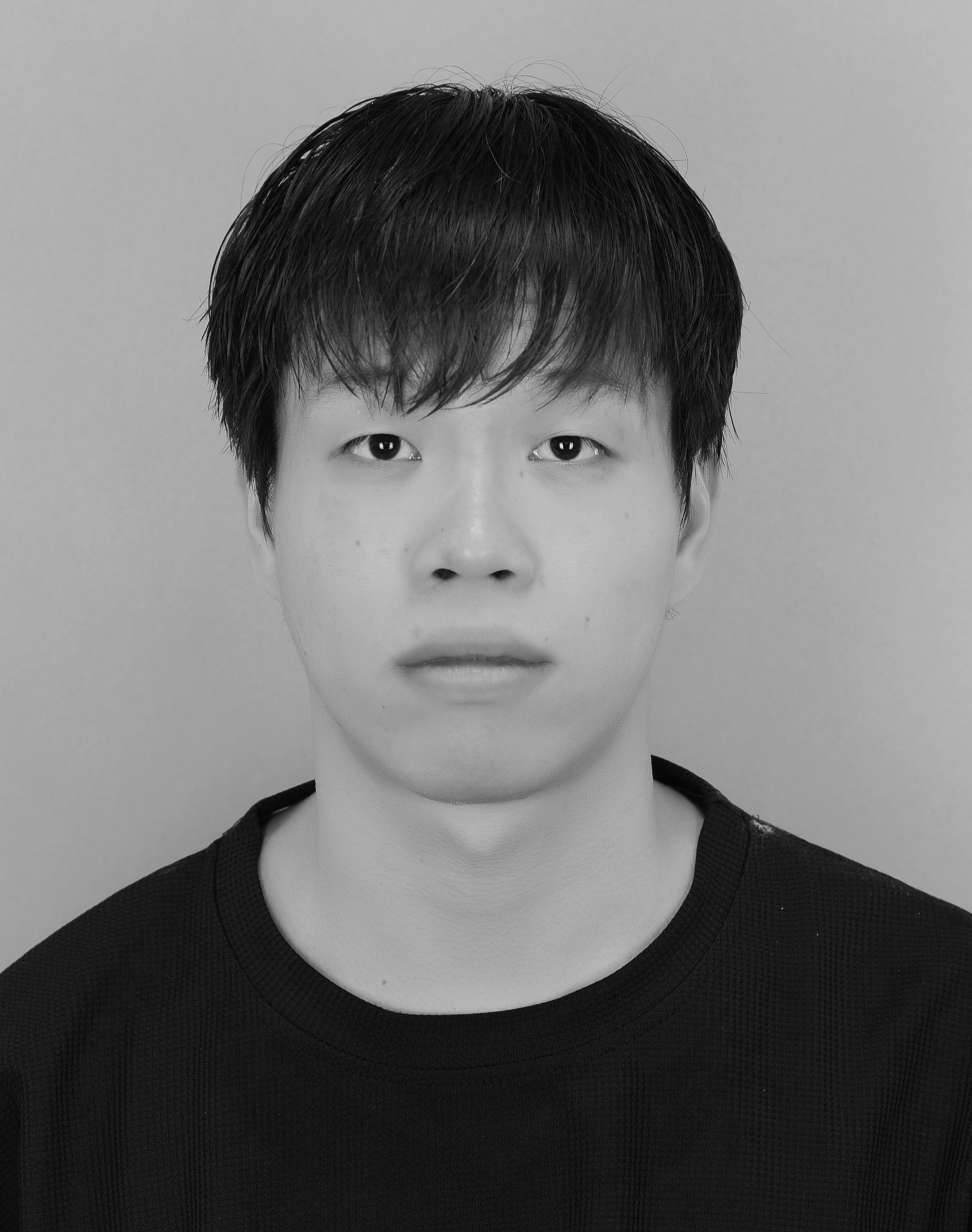}}]{Xin Yang}  received his bachelor's degree in Network Engineering from Guangdong University of Petrochemical Technology in 2019. He is currently pursuing a Ph.D. in Information and Communication Engineering at Shenzhen University. His research interests include robot navigation and autonomous driving. \end{IEEEbiography}
 
 \begin{IEEEbiography}[{\includegraphics[width=1in,height=1.25in,clip,keepaspectratio]{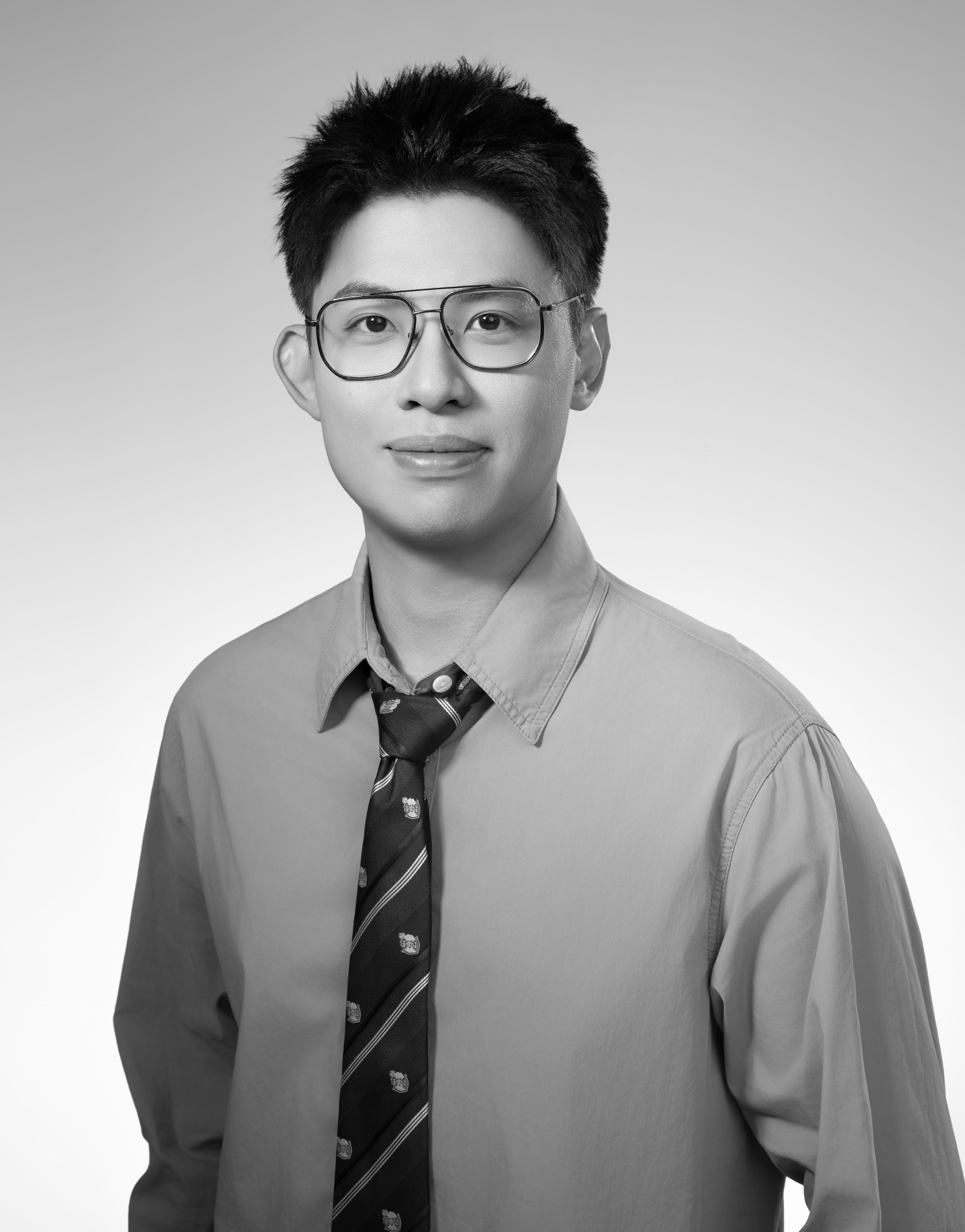}}]{Yuhang Zhang}
 	received the B.Sc. degree from Guangdong Ocean University, Guangdong, China, in 2019. He received the Ph.D. degree from Shenzhen University, Shenzhen, China, in 2024. From 2023 to 2024, he was a Researcher with the Institut d’Electronique et Technologies du numéRique of the VAADER Research Group, the National Institute of Applied Science (INSA Rennes), University of Rennes, France. He is currently a lecturer with the School of Computer Science and Cyber Engineering, Guangzhou University, Guangzhou, China. His research interests include computer vision, domain adaptation/generalization, semantic segmentation, and deep learning.
 \end{IEEEbiography}
\vspace{-5mm}
\begin{IEEEbiography}[{\includegraphics[width=1in,height=1.25in,clip,keepaspectratio]{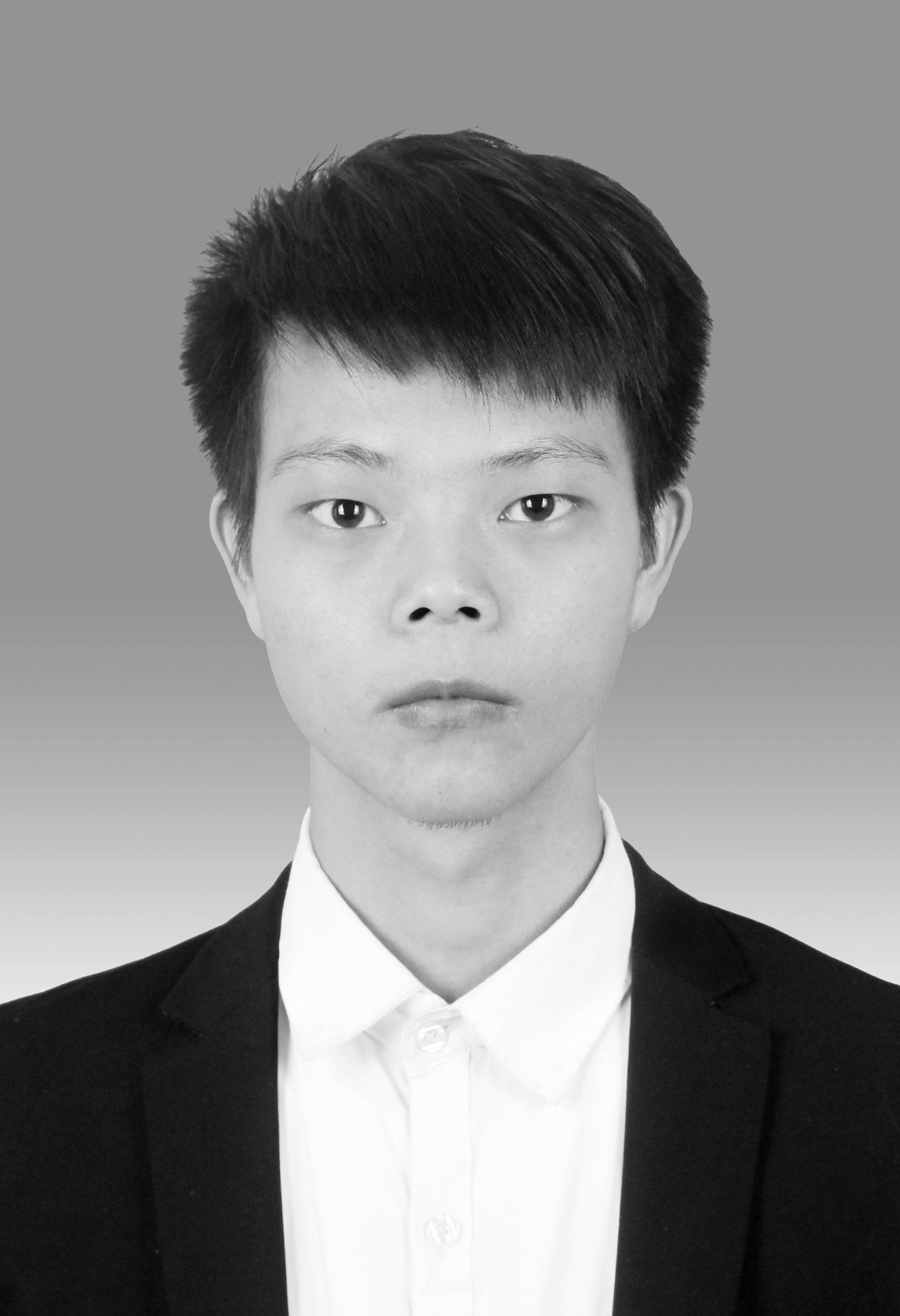}}]{Wei Li}
	received the B.Sc. and M.Sc. degrees from the Nanchang hangkong University, Jiangxi, China, in 2019 and 2022, respectively. He is currently pursuing the Ph.D. degree with the College of Electronics and Information Engineering, Shenzhen University, Shenzhen, China. His research interests focus on computer vision, image processing, and machine learning.
\end{IEEEbiography}
 \begin{IEEEbiography}[{\includegraphics[width=1in,height=1.25in,clip,keepaspectratio]{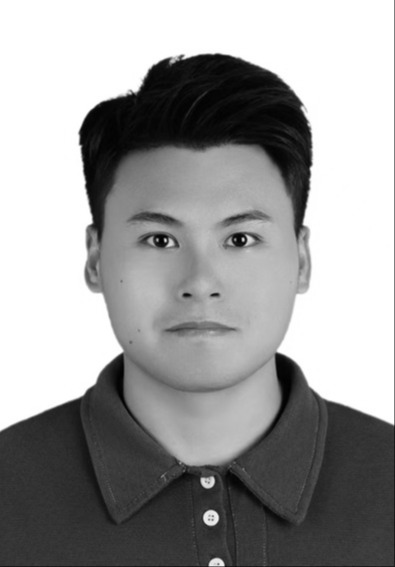}}]{Xin Lin }
 	received the Ph.D. degree from South China University of Technology, China, in 2022. He is currently an Assistant Professor with the School of Artificial Intelligence, Guangzhou University. He has authored or coauthored more than 20 articles in prominent journals and conferences, including IEEE Transactions on Image Processing, IEEE Transactions on Geoscience and Remote Sensing, IEEE Transactions on Circuits and Systems for Video Technology, CVPR, ICCV, AAAI, and ACM MM. His research interests include multi-modal large language models, visual comprehension, and embodied AI.
 \end{IEEEbiography}
 \begin{IEEEbiography}[{\includegraphics[width=1in,height=1.25in,clip,keepaspectratio]{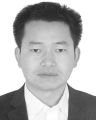}}]{Wenbin Zou}
 	received the M.E. degree in software engineering with a specialization in multimedia technology from Peking University, China, in 2010, and the Ph.D. degree from the National Institute of Applied Sciences, Rennes, France, in 2014. From 2014 to 2015, he was a Researcher with the UMR Laboratoire d’informatique Gaspard-Monge, CNRS, and Ecole des Ponts ParisTech, France.
 	\par In 2015, he joined Shenzhen University, Shenzhen, China, where he is currently an Associate Professor. His current research interests include saliency detection, object segmentation, and semantic segmentation.
 \end{IEEEbiography}

 \begin{IEEEbiography}[{\includegraphics[width=1in,height=1.25in,clip,keepaspectratio]{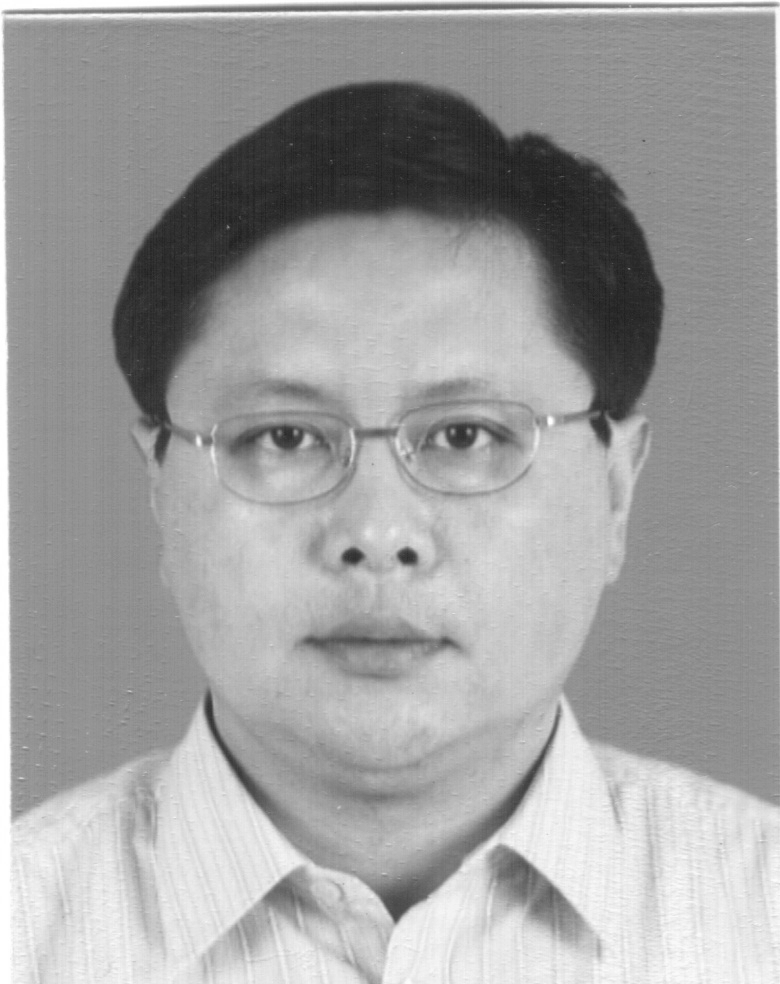}}]{Chen Xu}
 	received the B.Sc. and M.Sc. degrees from Xidian University, Xi an, China, in 1986 and 1989, respectively, and the Ph.D. degree from Xi an Jiaotong University, Xi an, in 1992.
 	\par In 1992, he joined Shenzhen University, Shenzhen, China, where he is currently a Professor. From 1999 to 2000, he was a Research Fellow with Kansai University, Suita, Japan, and the University of Hawaii, Honolulu, HI, USA, from 2002 to 2003. His current research interests include image processing, intelligent computing, and wavelet analysis.
 \end{IEEEbiography}

\end{document}